\documentclass[letterpaper, 10 pt, conference]{ieeeconf}  % Comment this line out if you need a4paper

\IEEEoverridecommandlockouts                              % This command is only needed if 
\usepackage{cite}
\usepackage{amsmath,amssymb,amsfonts}
\usepackage{graphicx}
\usepackage{textcomp}
\usepackage{xcolor}
\def\BibTeX{{\rm B\kern-.05em{\sc i\kern-.025em b}\kern-.08em
    T\kern-.1667em\lower.7ex\hbox{E}\kern-.125emX}}
\usepackage{times}
\usepackage{color,soul}
\usepackage[ruled,linesnumbered]{algorithm2e}
\usepackage{pgfgantt}
\usepackage{subfloat}
\usepackage{tabularx}
\usepackage{subfigure}
\usepackage[skip=0.01in,font=small,labelfont=bf]{caption}

\usepackage[colorlinks,bookmarksopen,bookmarksnumbered,citecolor=red,urlcolor=red]{hyperref}

\newcounter{myenumi}

\renewcommand\theenumi{\arabic{myenumi}. }

\long\def\Item#1\par{%
 \stepcounter{myenumi}%
 \makebox[1.5em]{\hl\theenumi}\hl{ #1}%
 \par
}

\renewcommand{\themyenumi}{
\setlength{\parindent}{0pt}% don't indent paragraphs
\arabic{myenumi}.}

\newenvironment{myenumerate}{%
\setlength{\parindent}{0pt}% don't indent paragraphs
\setcounter{myenumi}{0}% restart numbering
\renewcommand{\item}{% new definition of item
\par% start a new line
\refstepcounter{myenumi}% advance counter
\makebox[1.5em][l]{\themyenumi}% print counter to width of 3em, aligned to left
}% end of definition of item
}{% at end of environment
\par% start new paragraph
\noindent% don't indent new paragraph
\ignorespacesafterend% ignore spaces after environment
}

\title{\LARGE \bf
Risk-Aware Decision Making for Service Robots to Minimize Risk of Patient Falls in Hospitals}

\author{Roya Sabbagh Novin$^{1}$, Amir Yazdani$^{1}$, Andrew Merryweather$^{1}$, and Tucker Hermans$^{2}$% <-this % stops a space

\thanks{$^{1}$Department of Mechanical Engineering and Utah Robotics Center,
        University of Utah, Salt Lake City, Utah, USA, {\tt\small roya.sabbaghnovin@utah.edu}}
\thanks{$^{2}$School of Computing and Utah Robotics Center, University of Utah, Salt Lake City, Utah, USA}%
}
\begin{document}
\maketitle
\thispagestyle{empty}
\pagestyle{empty}

%%%%%%%%%%%%%%%%%%%%%%%%%%%%%%%%%%%%%%%%%%%%%%%%%%%%%%%%%%%%%%%%%%%%%%%%%%%%%%%%
\begin{abstract}
Planning under uncertainty is a crucial capability for autonomous systems to operate reliably in uncertain and dynamic environments. The concern of safety becomes even more critical in healthcare settings where robots interact with human patients. In this paper, we propose a novel risk-aware planning framework to minimize the risk of falls by providing a patient with an assistive device. Our approach combines learning-based prediction with model-based control to plan for the fall prevention task. This provides advantages compared to end-to-end learning methods in which the robot's performance is limited to specific scenarios, or purely model-based approaches that use relatively simple function approximators and are prone to high modeling errors. We compare various risk metrics and the results from simulated scenarios show that using the proposed cost function, the robot can plan interventions to avoid high fall score events.
\end{abstract}
%%%%%%%%%%%%%%%%%%%%%%%%%%%%%%%%%%%%%%%%%%%%%%%%%%%%%%%%%%%%%%%%%%%%%%%%%%%%%%%%
\section{INTRODUCTION}

% Motivation and Problem Statement

Recent advances in motion planning have significantly improved efficiency and safety in autonomous systems~\cite{schmerling2018multimodal, williams2018best}. One important topic in this area is decision-making under uncertainty in the presence of risk and constraints, specially, in mobile robot navigation scenarios and human-robot interaction problems~\cite{blackmore2011chance, nishimura2020risk}. Most of the literature on risk-aware planning consider the problem of collision avoidance in environments with dynamic obstacles and focus on minimizing the probability of collision to ensure safety~\cite{samuelson2018safety, chen2019crowd}. However, there are other types of risk that should be incorporated into planning for autonomous systems~\cite{medina2012risk}. An autonomous system should consider the risks associated with other agents during interactions, such as risk of musculoskeletal injury in human-robot interaction~\cite{yazdani2020estimating}, or the risk of failure of the other agent during throw-and-catch or handover applications~\cite{kober2012playing}.

A noteworthy application of risk-aware planning is assistive intelligence in healthcare~\cite{nejat2008can}. A service robot operating in populated environments needs to perceive and predict human movements, not only to avoid collision, but also to improve assistance in daily routines~\cite{mutlu2008robots, bennewitz2002learning}. A significant burden in health care occurs because of high patient-to-nurse ratios~\cite{alert2015preventing}, contributing to an unacceptably high rate of patient falls. Patient sitters may not be cost effective, but have been seen as an effective fall prevention intervention~\cite{chu2017preventing}. Previously, we proposed the use of an autonomous mobile robot to actively respond to a patient when unstable by providing an assistive mobility aid~\cite{sabbagh2021model}. However, we assumed the patient's pose to be known and stationary. In this paper, we introduce the problem of risk-aware planning for an autonomous mobile robot to minimize the risk of falls while intervening to assist patients moving to uncertain goals. 

\begin{figure}[t!]
\centering
\includegraphics[width = 7.5cm]{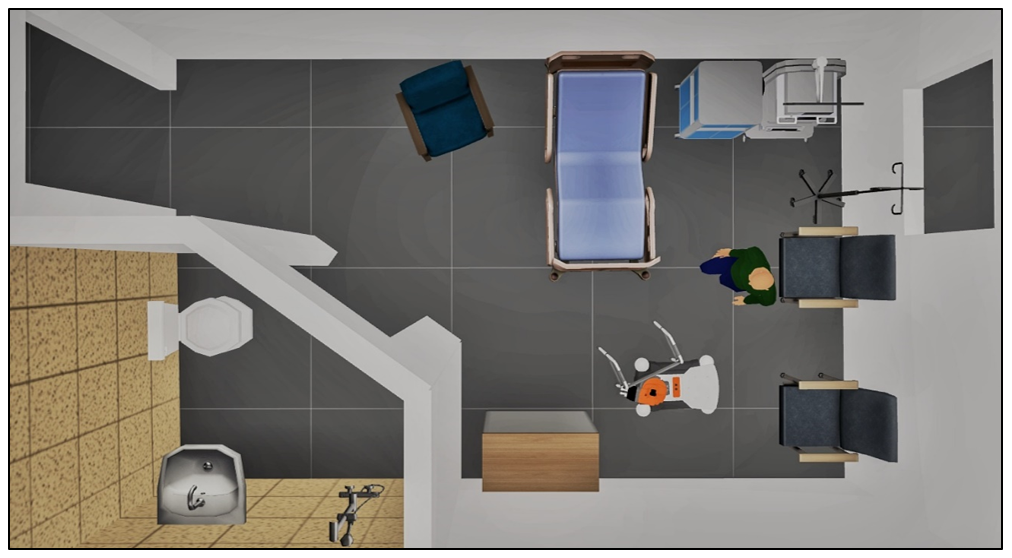}
\caption{\small{Our simulated hospital room environment in Gazebo, in which the service robot decides when and where to deliver a walker to the patient, considering the risk of patient fall.}}
\label{fig:setup}
\vspace{-0.3in}
%\rule{3cm}{7cm}
\end{figure}

% Challenges & Prior Work
An important component in risk-aware planning is human motion prediction which aims to forecast the future behavior of a human based on the observed trajectory up until the current time. Predicting human motion is a well-studied topic, especially in the context of pedestrian trajectory prediction~\cite{ziebart2009planning,fajen2003dynamical}. Learning-based methods have shown excellent prediction with reduced computation time~\cite{monfort2015intent}. In this work, motivated by successful methods in the autonomous vehicle literature~\cite{schmerling2018multimodal}, we combine learning-based patient trajectory prediction with model-based control to plan for an assistive robot that provides support through a mobility aid. We investigate various risk metrics to find which one is the most successful in avoiding high-risk situations. 

% Contributions
Our key contributions can be summarized as follows. First, we propose a risk-aware planning framework that considers the risk of patient falls in the decision making process to find the optimal location and time to provide a supporting mobility aid. We use a novel risk-aware cost function for stochastic optimization under uncertainty leveraging the conditional value at risk (CVaR) metric. Our cost function penalizes even low probability events associated with high fall score in order to fulfill the goal of patient fall prevention. To evaluate fall scores within the room, we use the fall risk assessment tool developed in~\cite{novin2020development} which can provide the fall score distribution over the room layout and along a trajectory.

Second, we develop a simulated patient model in Gazebo simulator to avoid complications of human subjects at high risk of injury from a fall. Fig.~\ref{fig:setup} presents our simulation setup that provides realistic trajectories and animations of human motion with the capability of adding new motion from collected data. It also provides probabilistic predictions over patient intention and trajectory based on learned models.

% We organize the remainder of the paper as follows. We briefly discuss prior work in human motion prediction and risk-aware planning in Section~\ref{sec:prior_work}. In Section~\ref{sec:methods}, we introduce and formalize the optimization problem to minimize the risk of fall in hospitals. We follow this with details of our implementation of human motion and fall score prediction and optimization for risk-aware planning in Section~\ref{sec:implementation}. We analyze the results from simulation experiments in Section~\ref{sec:evaluation}. Finally, we present a closing discussion including limitations and potential future work in Section~\ref{sec:conclusion}.

\section{Prior Work}
\label{sec:prior_work}
Planning appropriate robot motion in scenarios where both the human and the robot perform tasks in a shared workspace is a well-studied, but still challenging, problem due to the uncertain nature of human behavior~\cite{majumdar2020should}. In such scenarios, risk-aware motion planning mostly refers to planning robot motions to avoid collisions with humans while keeping the path smooth~\cite{park2019planner,bandyopadhyay2013intention}. The primary approaches to human motion prediction include analytical methods~\cite{fajen2003dynamical}, optimization-based methods~\cite{srinivasan2006computer, ackermann2010optimality}, and inverse reinforcement learning~\cite{ziebart2009planning, monfort2015intent}. Learning-based methods use relevant features affecting human motion to incorporate predictive models in motion planning~\cite{kretzschmar2016socially, karasev2016intent}. Trautman~et~al.\ use a Gaussian process for cooperative navigation for a mobile robot through dense human crowds~\cite{trautman2015robot}.

Multi-policy decision making is another framework for navigation in dynamic environments under uncertainty, switching between a set of policies evaluated using forward simulations. Mehta et al.\ contend with the complexity of policy evaluation by representing a forward simulation as a differentiable deep network and enabling effective backpropagation~\cite{mehta2018backprop}. Grey et al.\ propose a system in~\cite{grey2016humanoid} for solving humanoid manipulation problems that uses a hybrid backward-forward planning algorithm as a task planner along with humanoid manipulation primitives. Fisac~et~al.\ use the concept of confidence in the learned human motion model for robot planning~\cite{fisac2018probabilistically}. Particularly, their method leverages the rationality coefficient in the human model as an indicator of the model’s confidence. They combine this confidence-aware human motion predictions with a safe motion planner to obtain probabilistically-safe robotic motion plans.

A set of approaches use probabilistic distributions to represent uncertainty in the system~\cite{van2012motion, mukadam2018continuous, schmerling2018multimodal, williams2018best}. Recent methods for dynamic collision avoidance include iterative local optimization in belief space~\cite{van2012motion}, Gaussian process motion planning~\cite{mukadam2018continuous}, exhaustive search with motion primitives~\cite{schmerling2018multimodal}, and information-theoretic control~\cite{williams2018best}. Although these approaches find an optimal solution, they only optimize for the expected cost and are considered risk-neutral.

There are other methods that have introduced risk metrics to provide risk-awareness in the planning scheme~\cite{blackmore2011chance,samuelson2018safety,medina2012risk}. Nishimura et al.\ present a risk-sensitive stochastic optimal control framework for safe crowd-robot interaction~\cite{nishimura2020risk}. They model the risk by an entropic risk measure and incorporate it into a stochastic controller. Hakobyan et al. propose a risk-aware motion planning and decision-making method in uncertain and dynamic environments~\cite{hakobyan2019risk}. They develope a two-stage strategy consisting of (i) generating a safe reference trajectory using RRT* and (ii) utilizing CVaR to assess safety risks and design a CVaR risk-constrained receding horizon controller to track the reference trajectory. Singh et al.\ \cite{singh2018risk} propose a risk-sensitive inverse reinforcement learning approach that accounts for risk sensitivity in humans. They infer human’s underlying risk preferences from demonstrations based on a broad range of coherent risk metrics. They use their framework to capture different driving styles, ranging from risk-neutral to highly risk-averse.

All the methods mentioned above only consider risk in terms of collision avoidance. However, there are other types of risks, especially in environments where humans exist that if considered, can improve a robot's performance. To address these type of risks, we must define appropriate risk metrics that consider the uncertainty associated with various events. Therefore, in this work, we propose a framework to include these risk metrics in robot planning. We use non-parametric probabilistic Gaussian processes (GPs) for human intention and motion prediction and use the resulting fall score distribution as the loss function in the planning optimization. For risk-aware planning, we employ a combination of the expected cost and CVaR risk metric in a stochastic optimization. We believe having both of these components in the objective function (rather than as constraints) not only minimizes the overall risk of fall, but also minimizes the probability of tail events in which safety losses exceed a user-specified amount defined as value at risk (VaR). 
%%%%%%%%%%%%%%%%%%%%%%%%%%%%%%%%%%%%%%%%%%%%%%%%%%%
\section{Methods}
\label{sec:methods}
In this section we introduce a framework for manipulation planning under uncertainty considering patient fall risk criteria. In our framework, the goal is to perform the task and manipulation planning while considering a risk model. Specifically, for the application of fall prevention, we want to find the best manipulation plan to provide the patient with better support and minimize the risk of fall. We use a fall risk assessment model developed in~\cite{novin2020development} to find the fall score distribution.\footnote{We use ``fall score'' instead of ``fall risk'' to avoid any confusion since we use the term ``risk'' to denote the economic concept underlying CVaR.} In the following, we introduce the high-level problem of fall prevention optimization and briefly discuss our method for human motion and intention prediction. We then review various risk metrics for use as cost functions and provide details of our proposed risk-aware planning optimization.

% \begin{figure}[t!]
% \vspace{0.1in}
% \centering
% \includegraphics[width = 8cm]{Figures/framework.pdf}
% \caption{\small{The planning framework to minimize risk of patient falls.}}
% \label{fig:framework}
% \vspace{-0.2in}
% %\rule{3cm}{7cm}
% \end{figure}

\subsection{Problem statement}

\begin{figure}[t!]
\vspace{0.05in}
\centering
\includegraphics[width = 5.7cm, height = 4.2cm]{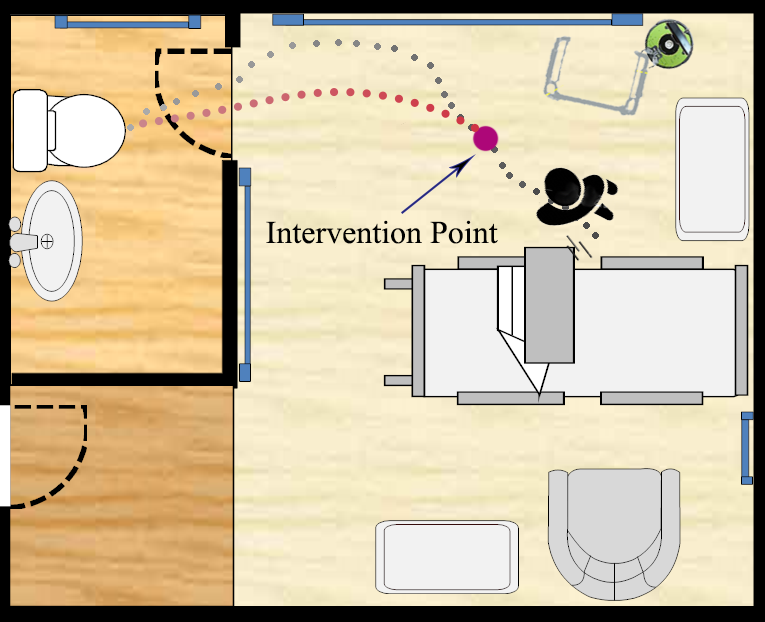}
\caption{\small{The intervention point is defined as the location where the robot gives the supporting object to the patient. Here, the gray dotted line shows the predicted patient trajectory without having the supporting object. Once the patient gets the supporting object the predicted trajectory changes to the red dotted one.}}
\label{fig:intervention_point}
\vspace{-0.2in}
%\rule{3cm}{7cm}
\end{figure}

Assuming that we have an assistive robot that can manipulate objects within the room similar to the one proposed in~\cite{novin2018dynamic}, the robot still needs to decide where and when to deliver the object. Hence, the main challenge in fall prevention using an additional supporting object is to find where the optimal location is to put the supporting object and when it should be there. We refer to this location as the ``\textit{intervention point}'' and define the problem as finding the best intervention point and manipulation plan that minimizes the fall risk probability with the minimum effort by the robot (Fig.~\ref{fig:intervention_point}). Formally, we define our optimization as:
\begin{flalign}
\underset{I, \boldsymbol{\zeta}_I}{\text{min}}&\quad
J = [(1-\alpha_t) c_F(\boldsymbol{\tau}_P)+\alpha_t c_F'(\boldsymbol{\tau}_P')]+\rho[c_M(\boldsymbol{\tau}_R)]\nonumber\\
\text{s.t.}& \quad \alpha_t = 0, \quad\forall t = 0, \dots, I\nonumber\\
&\quad \alpha_t= 1, \quad \forall t = I, \dots, T\nonumber\\
&\quad ||\boldsymbol{\zeta}_I^R-\boldsymbol{\zeta}_I^P||_2^2\leq \epsilon\label{eq:optimization}
\end{flalign}
The decision variables $I$ and $\boldsymbol{\zeta}_I$ define the point of intervention and time of intervention.
The objective function of our problem contains two parts. The first includes the patient fall risk and the second part encodes the cost of manipulation. $c_F(\boldsymbol{\tau}_P)$ and $c_F'(\boldsymbol{\tau}_P')$ define the risk of fall before and after the intervention which can be obtained using the fall risk assessment model in~\cite{novin2020development} for a given trajectory. We use $\alpha$ to indicate which cost (unaided vs aided) is used for each part of the patient trajectory. Moreover, there exists a trade-off between minimizing the risk of fall ($c_F(\boldsymbol{\tau}_P)$) and the cost of manipulation ($c_M(\boldsymbol{\tau}_R)$) which we assign by a weight $\rho$. We use a small value for $\rho$ that prioritizes minimizing fall risk as the main objective while ensuring feasibility of the robot's motion plan. We use the manipulation planning framework introduced in~\cite{novin2019model} to calculate the manipulation cost. 

The last line of constraints insures the reachability of the provided supporting object by the patient. We assume that the patient can only grab the walker if it is in a reachable distance (defined by $\epsilon$), indicating a relation between the robot's state $\boldsymbol{\zeta}_I^R$ and the patient's state $\boldsymbol{\zeta}_I^P$ at the time of intervention.
Other constraints related to the patient motion and the robot path appear inside their respective trajectories ($\boldsymbol{\tau}_P, \boldsymbol{\tau}_R$). 

Since the trajectory optimization problem is continuous and uncertain, we can naturally view the associated risk-aware planning problem from the perspective of probabilistic inference~\cite{botvinick2012planning}. We introduce a binary variable $\mathcal{O}$ for optimality, where $\mathcal{O}=1$ when the patient's motion and the robot's~plan are optimal and $\mathcal{O}=0$ when they are suboptimal. Given the intervention point, we can say the patient's optimality and the optimality of the robot motion are conditionally independent.
\begin{gather} 
\begin{split}
 p(\mathcal{O}=1|\boldsymbol{\zeta}_I) &= p(\mathcal{O}_P=1,\mathcal{O}_R=1|\boldsymbol{\zeta}_I)\\
 &=  p(\mathcal{O}_P=1|\boldsymbol{\zeta}_I)\cdot p(\mathcal{O}_R=1|\boldsymbol{\zeta}_I)
\end{split}
\end{gather} 

Our goal now is to solve the following optimization
\begin{gather}
I^*,\boldsymbol{\zeta}_I^* = \arg \max_{I,\boldsymbol{\zeta}_I} \{p(\mathcal{O}=1|\boldsymbol{\zeta}_I)\}
\label{eq:probabilistic_optimization}
\end{gather} 
to find the time and location of intervention to maximize the probability of optimality. To solve this optimization, first, we need to define the optimality probability for each agent. We define the robot's optimality model in a straightforward way as the exponential of the effort required to follow its path. We use the method presented in~\cite{novin2019model} for mobile manipulation planning and use the cost provided in that approach ($c_M$) as the robot's effort:
\begin{gather}
p(\mathcal{O}_R=1|\boldsymbol{\zeta}_I) = \exp(\rho c_M(\boldsymbol{\tau}_R))
\label{eq:robot_optimality}
\end{gather}
Here, we define $\rho$, similar to Eq.~(\ref{eq:optimization}), as a parameter to adjust the importance of robot cost against the patient's safety.

For the patient's motion, we define optimality based on the risk of fall. We use Gaussian processes (GP) to represent the human motion within the room~\cite{das2018cross}. We need to define an appropriate risk metric to find the overall fall score given possible scenarios which is a predicted distribution over trajectories.

\subsection{Human motion and intention prediction}
\label{sec:human}
Here, we briefly review Gaussian process progression (more details can be found in~\cite{williams2006gaussian}) and provide an overview of our human motion prediction scheme. 

We implement our human motion model as a GP, with input as the current state $\boldsymbol{\zeta}_t^P \in \mathbb{R}^2$ and output as the change in the next step $\boldsymbol{\Delta}_t=\boldsymbol{\zeta}^P_{t+1}-\boldsymbol{\zeta}^P_t$. The GP model is:
\begin{gather}
    \boldsymbol{\Delta}_i= f(\boldsymbol{\zeta}^P_i) + \epsilon_i,\quad f\sim GP(0,\mathcal{K}),\; \epsilon_i \sim \mathcal{N}(0,\sigma^2)
    \label{eq:GP model}
\end{gather}
% \setlength{\textfloatsep}{2pt}
% \begin{subfigures}
% \begin{figure}
% \vspace{-0.32cm}
% \end{figure}
% \begin{algorithm}[!t]
% \DontPrintSemicolon
% \SetAlgoLined
% \SetNlSty{texttttt}{(}{)}
% \SetKwInOut{Input}{input}
% \SetKwInOut{Output}{output}
% \Input{Model $f_g$, time horizon $T$, number of sample trajectories $K$} 
% Initialize $\boldsymbol{\zeta}_0^P=$ current state\;
% \For { $k=1$ $\mathrm{to}$ $K$}{
% \For { $t=0$ $\mathrm{to}$ $T-1$}{
% $\boldsymbol{\zeta}_{t+1}^P$ $\sim$ $p(\boldsymbol{\zeta}_{t+1}^P|\boldsymbol{\zeta}_{t}^P, g)$ \;
% Add $\boldsymbol{\zeta}^P_{t+1}$ to $\boldsymbol{\tau}_P^k$
% }
%     }
% \caption{Predicting trajectory distribution for a given GP model.}
% \label{alg:propagation}
% \end{algorithm}
% \end{subfigures}
The prior on $f$ is a Gaussian process, as a result $p(f)$ and the likelihood $p(\boldsymbol{\Delta}_i|f)$ are both Gaussian as well. GPs are defined by a mean function $\mu(\boldsymbol{\zeta}^P)$, and a covariance kernel function $\Sigma_{i,j}=\mathcal{K}(\boldsymbol{\zeta}^P_i,\boldsymbol{\zeta}^P_j)$. We consider a prior mean function $\mu(\boldsymbol{\zeta}^P)=0$ and a squared exponential kernel function:
\begin{gather}
    \mathcal{K}(\boldsymbol{\zeta}^P_i,\boldsymbol{\zeta}^P_j) = \alpha^2 \exp\left(-\frac{1}{2}(\boldsymbol{\zeta}^P_i-\boldsymbol{\zeta}^P_j)^T\Gamma^{-1}(\boldsymbol{\zeta}^P_i-\boldsymbol{\zeta}^P_j)\right)
\end{gather}
where $\alpha^2$ defines the variance of function $f$ and $\Gamma$ is the diagonal matrix of length-scales. GPs are fit to the training data ($\mathbf{Z},\boldsymbol{\Delta}$) by optimizing the evidence to obtain posterior hyperparameters. After the hyperparameters are tuned, the one-step prediction distribution of the output for a test input $\mathbf{x}_*$ can be obtained by:
\begin{gather}
    p(\boldsymbol{\Delta}_{*}|\boldsymbol{\zeta}^P_*) = \mathcal{N}(\mu_*,\sigma_*^2)\\
    \mu_*=\Sigma_*^T(\Sigma+\sigma^2I)^{-1}\Delta\\
    \sigma_*^2 = \Sigma_{**} - \Sigma_*^T(\Sigma+\sigma^2\mathbf{I})^-1\Sigma_*
\end{gather}
where $\Sigma_* = \mathcal{K}(\boldsymbol{\zeta}^P,\boldsymbol{\zeta}^P_*)$, and $\Sigma_{**} = \mathcal{K}(\boldsymbol{\zeta}^P_*,\boldsymbol{\zeta}^P_*)$. Training time for GPs can be long for large datasets. Prediction, however, is fast, and hence our primary motivation for using GPs is the real-time risk estimation performance.

In our problem, we use a mixture of GPs to cover all the possible scenarios within the room. Following~\cite{das2018cross}, for each intent $g$ (i.e. goal location), we use an atomistic approach in which we learn a separate model for each coordinate in the 2D space ($p(x_{t+1}|\boldsymbol{\zeta}^P_t,g)$, $p(y_{t+1}|\boldsymbol{\zeta}^P_t,g)$) and use the joint distribution with independence assumptions to obtain:
\begin{gather}
    p(\boldsymbol{\zeta}^P_{t+1}|\boldsymbol{\zeta}^P_t,g) = p(x_{t+1}|\boldsymbol{\zeta}^P_t,g)p(y_{t+1}|\boldsymbol{\zeta}^P_t,g)
    \label{eq:one_step}
\end{gather}

After learning the GP mixture model, we can use it to predict both intention and future trajectories given an intention. We use Bayes rule and the first-order Markov assumption to find the intent distribution as:
\begin{gather}
    p(g|\boldsymbol{\zeta}^P_{1:t+1}) \propto p(\boldsymbol{\zeta}^P_{t+1}|\boldsymbol{\zeta}^P_{t},g)p(g|\boldsymbol{\zeta}^P_{1:t})^{1-\lambda}
    \label{eq:pred}
\end{gather}

The first term on the right is calculated using Eq.~(\ref{eq:one_step}) and the second term is obtained by recursive evaluation from the initial belief over intentions $p(g|\boldsymbol{\zeta}^P_1$) which is predefined. We add another parameter as the forgetting factor $0<\lambda<1$, which indicates the amount of past observations to forget. 

For trajectory prediction, we sample $K$ random trajectories from the GP models. At each time step $t$ for each trajectory $k$, we use the one-step prediction from Eq.~(\ref{eq:one_step}) to find the distribution for the next time step. We sample a point from that distribution and add it to trajectory $k$.

\begin{figure}[t!]
\vspace{0.05in}
\centering
\includegraphics[scale = 0.41]{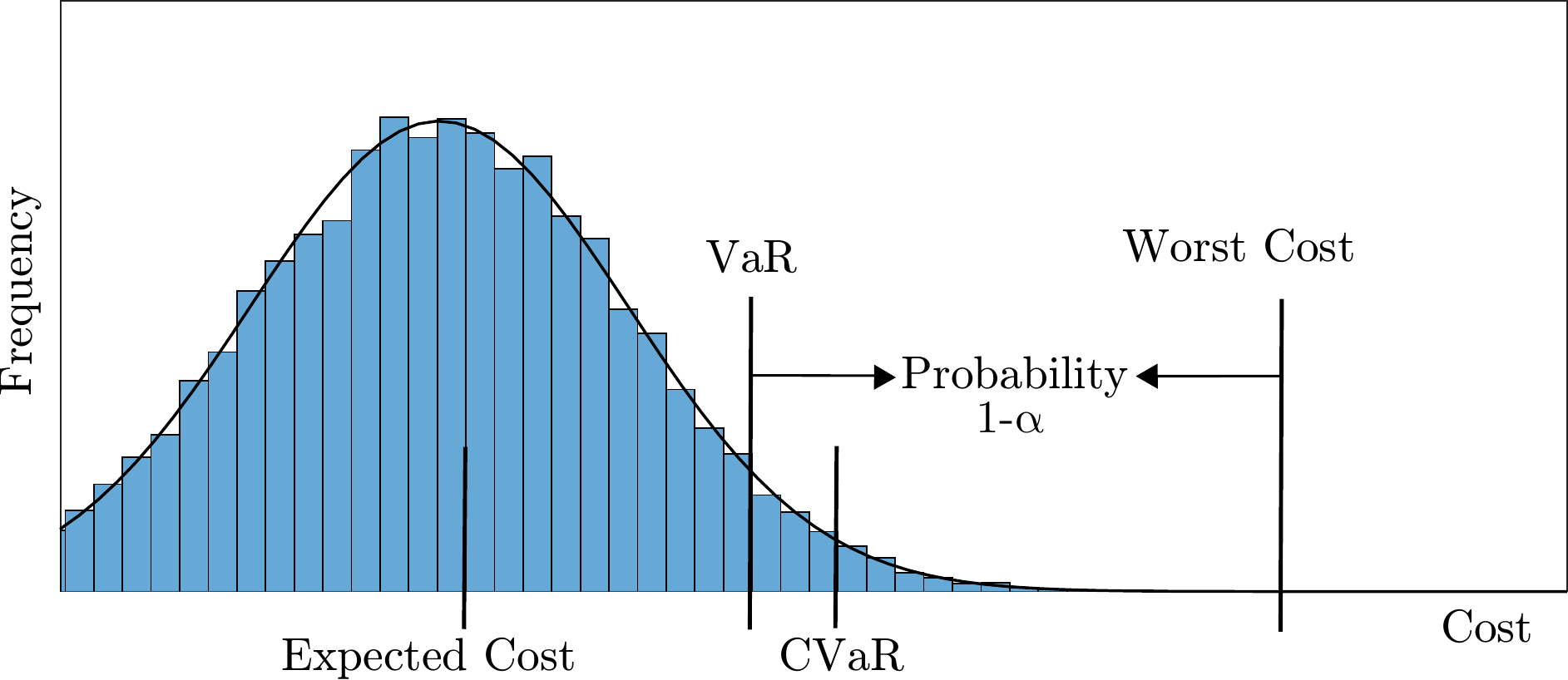}
\caption{\small{Different risk metrics (figure inspired by~\cite{majumdar2020should}).}}
\label{risk}
\vspace{-0.2in}
%\rule{3cm}{7cm}
\end{figure}

\subsection{Risk metrics}
\label{sec:risk}
Fig.~\ref{risk} provides a visualization for different risk metrics that are used in the literature on risk-aware decision making. Here, ``\emph{risk}" refers to the possibility of having high cost~(i.e. high fall score). The most common stochastic planning risk metric used in robotics is the expected value of a cost function. However, in applications where we need to account for risk, the expected value is not the best choice since it is neutral to the risk.

An alternative approach is to consider worst-case assessment of the distribution of the stochastic cost function~\cite{majumdar2020should}. This approach can be too conservative and probably overreact to simple situations. In our case, it might result in advising the patient to never leave the bed, or bringing the walker to the bedside every time the patient is getting out of the bed, which is inefficient and likely annoying for the patient. Responding to all situations, even those with low fall scores, can drain the robot's battery and make it unable to respond to a future high fall risk situation. For practical applications, such as fall prevention, we need a risk metric that lies in between these extremes. A better risk metric, mostly used in finance, is the conditional value at risk ($\mathrm{CVaR_\alpha}$) defined as:
\begin{gather}
\mathrm{CVaR}_{\alpha}(Z) :=\frac{1}{\alpha}\int_{1-\alpha}^1 \mathrm{VaR}_{1-s}(Z)ds
\end{gather}
where $\mathrm{VaR_\alpha}$ is the Value at Risk at level $\alpha$:
\begin{gather}
\mathrm{VaR}_{\alpha}(Z) := \min\{z|P[Z>z]\leq\alpha\}
\end{gather}

Intuitively, $\mathrm{CVaR_\alpha}$ indicates the expected value of $Z$ for the upper $(1-\alpha)$-tail of the $Z$ distribution. We propose using a linear combination of expected value and $\mathrm{CVaR_\alpha}$ in order to minimize the overall fall score while avoiding even low probability events with high fall scores:
\begin{gather}
p(\mathcal{O}_P=1|\boldsymbol{\tau}_P) =  \exp\bigg(-\mathrm{E}[c_F(\boldsymbol{\tau}_P)]- \beta \mathrm{CVaR_\alpha}[c_F(\boldsymbol{\tau}_P)]\bigg)
\end{gather}
In the above, $\boldsymbol{\tau}_p$ is the patient's motion and $c_F$ is the fall score function based on the patient motion.

\subsection{Risk-aware planning}
Our objective is to find the optimal intervention point (pose and time) for the robot to minimize the patient's fall score given a mixture of GPs for patient motion prediction and a patient fall score model. Given the patient motion observed up to the current time $\boldsymbol{\tau}_P^-$, we compute the probability of optimality as the sum of optimality over all the possible future trajectories for $m$ number of intentions:
\begin{gather}
p(\mathcal{O}_P=1|\boldsymbol{\zeta}_I) = \sum_{j=1}^m p(\mathcal{O}_P=1|\boldsymbol{\tau}_P^+, \boldsymbol{\zeta}_I)p(\boldsymbol{\tau}_P^+|g_j)p(g_j|\boldsymbol{\tau}_P^-)
\end{gather}
where $\boldsymbol{\tau}_P^+$ is the predicted patient's motion. The second and third terms on the right are calculated using Eqs. (\ref{eq:one_step}) and (\ref{eq:pred}). Using the proposed risk metric, we find the optimality for a given trajectory and intervention point as a combination of the fall score before intervention~(without support) and after intervention~(with support).
\begin{gather}
\begin{split}
p(&\mathcal{O}_P=1|\boldsymbol{\tau}_P^+,\boldsymbol{\zeta}_I)\\
&= \sum_{t=0}^I \exp\bigg(-\mathrm{E}[c_F(\boldsymbol{\tau}_P)]- \beta \mathrm{CVaR_\alpha}[c_F(\boldsymbol{\tau}_P)]\bigg)\\
&+\sum_{t=I}^T \exp\bigg(-\mathrm{E}[c_F'(\boldsymbol{\tau}_P')]- \beta \mathrm{CVaR_\alpha}[c_F'(\boldsymbol{\tau}_P')]\bigg)
\end{split}
\label{eq:patient_optimality}
\end{gather}
where $c_F'$ and $c_F$ represent fall models with and without external support, respectively, and  $\boldsymbol{\tau}_p'$ is the future trajectory distribution assuming that the patient has a mobility aid. $\boldsymbol{\tau}_p'$ is computed the same way as $\boldsymbol{\tau}_p$ but with a different trajectory generation model.

Finally, by substituting Eqs. (\ref{eq:robot_optimality}) and (\ref{eq:patient_optimality}) in Eq.~(\ref{eq:probabilistic_optimization}), we can find the optimal intervention location based on the patient fall risk and the robot manipulation cost. We compare methods to solve this optimization in section~\ref{sec:opt}.
%%%%%%%%%%%%%%%%%%%%%%%%%%%%%%%%%%%%%%%%%%%%%%%%%%%%%%%%
\section{Implementation}
\label{sec:implementation}
In the following, we provide more details on the implementation aspects of our proposed framework. We discuss our simulated patient model, and our optimization methods and structures to find the optimal plan for the robot.
\subsection{Simulated patient model}
We begin our implementation by simulating patient trajectories within the room. Due to privacy limitations, we do not currently have access to real data collected in hospitals. We instead generate semi-realistic trajectories using an optimization-based approach. We note that once we acquire real hospital data, we can simply train our models on that data without any other changes to our proposed framework.

In trajectory simulation, our main assumption is that humans try to be optimal in their actions. We also assume that hospital patients are fragile and try to move as close as possible to supporting objects along their path. Based on these assumptions, we define an optimization problem with two main costs: (1) the length of trajectory, and (2) distance to the nearest supporting object~(please see the supplementary document available at \href{https://sites.google.com/view/risk-aware-decision-making}{https://sites.google.com/view/risk-aware-decision-making} for more details on patient trajectory generation).

Once we have the human motion dataset, we train the GP models as discussed in Section \ref{sec:human}. Using the GP mixture model we can predict both human intentions and trajectories. Finally, we take the predicted trajectories as the input to the fall score evaluation model introduced in~\cite{novin2020development} to obtain the fall scores along each predicted trajectory.

As an additional contribution of this paper, we release an open-source plugin for patient model in the Gazebo simulator, as shown in Fig.~\ref{fig:setup}. The simulated patient motion and intention is sampled from the GP trajectory distribution. The type of activity is determined based on the sampled intention and distance to the objects. The simulated patient can perform common activities such as ``sit-to-stand", ``stand-to-sit", and ``walking". However, more activities can be added using motion data collected from human subjects.

\subsection{Decision making optimization}
\label{sec:opt}

As a baseline for our experiments, we use a deterministic search-based optimization. First, we find the most probable human intention and obtain predicted trajectory for that intention. Then, we search along the predicted trajectory to find the optimal pose and time for intervention. The cost used in this optimization is the same as Eq.~(\ref{eq:optimization}).

As an alternative approach, we propose solving the probabilistic optimization problem. For this, we employ the cross-entropy method (CEM) which is a sample-based optimization technique commonly used for motion planning~\cite{de2005tutorial, kroese2006cross}. CEM is an iterative algorithm with two key steps~\cite{kobilarov2012cross}:
\begin{myenumerate}
    \item Generate random samples according to a specified probability distribution function.
    \item Update the parameters of the probability distribution using the ``elite" set of samples, denoted by the cost function, to produce better samples in the next iteration.
\end{myenumerate}

\begin{figure}[t!]
\vspace{0.05in}
\centering
\includegraphics[width = 8.5cm]{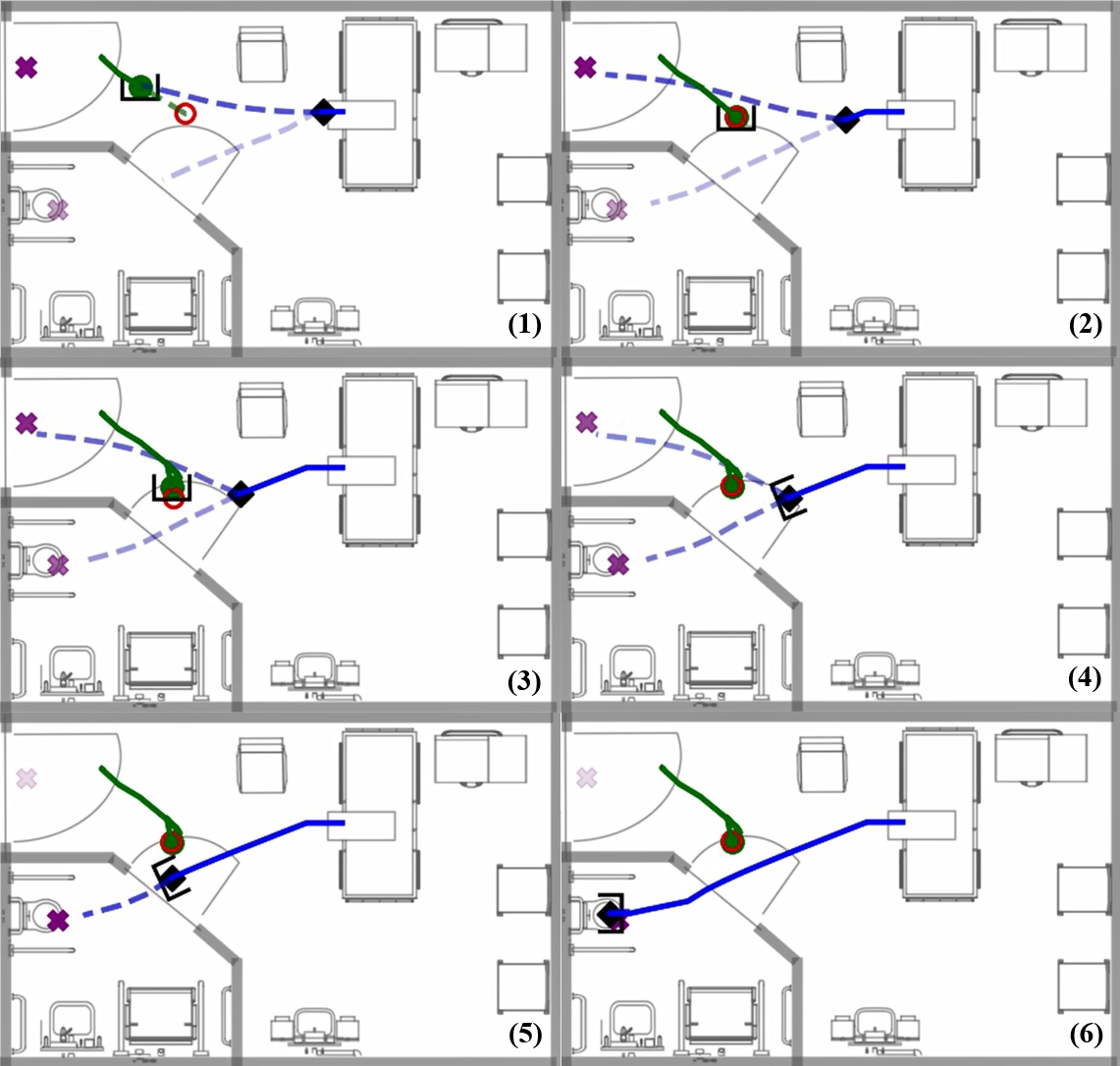}
\caption{\small{Overview of our risk-aware planning framework. Solid blue line: the observed patient's trajectory; dashed blue line: trajectory predictions given probable goals; purple X markers: the probable goals; solid green line: robot's trajectory; dashed green line: robot's plan; red circle: best intervention point. The opacity level of the predicted trajectories and intentions represent their probability.}}
\label{fig:prediction}
\vspace{-0.2in}
%\rule{3cm}{7cm}
\end{figure}

The above steps are repeated until the set of samples converge based on the Kullback–Leibler (KL) distance between the previous distribution and the new distribution. In addition to specifying the family of sampling distributions, the CEM algorithm depends on several user-defined parameters, including the elite fraction ($\gamma=0.1$) which is the ratio of number of elite samples to the total number of samples, the total number of samples ($N=100$), and the convergence criteria ($\epsilon=0.2$). We also define the initial sample distribution $(\mu_0, \sigma_0)$ in such a way that it covers the entire room space.

We explore different cost functions using various combinations of risk metrics as mentioned in Section \ref{sec:risk} and compare the resulting fall score distributions to find the most effective one. In the following section, we discuss the results from these experiments.

%%%%%%%%%%%%%%%%%%%%%%%%%%%%%%%%%%%%%%%%%%%%%%%%%%%
\section{Evaluation}
\label{sec:evaluation}
In this section, we provide results from our experiments. We compare the fall risk distribution with and without the robot's intervention, as well as various risk metrics discussed in this paper. We also compare our probabilistic approach with a deterministic solution. 

Fig.~\ref{fig:prediction} shows an overview of our risk-aware planning framework for one scenario, in which the initial pose of the human is known. In this figure, the solid blue line represents the observed trajectory from the patient, the dashed blue lines are trajectory predictions given the probable goals and the purple ``X" markers show the probable goals. The opacity level of the predicted trajectories and intentions represent their probability. At each time step, the robots gets an observation of the human pose and updates the predicted trajectories. Based on the new predictive distribution, it finds the best intervention point~(red circle) and executes the plan until the next observation. Once the walker is close enough to the patient, the patient grabs it and continues their path with the walker.

\begin{figure}[t!]
\centering
% \begin{subfigure}
%  \centering
%  \includegraphics[width = 7.2cm]{Figures/fall_distribution_bed.pdf}
% \end{subfigure}
\begin{subfigure}
 \centering
\includegraphics[width = 7.1cm]{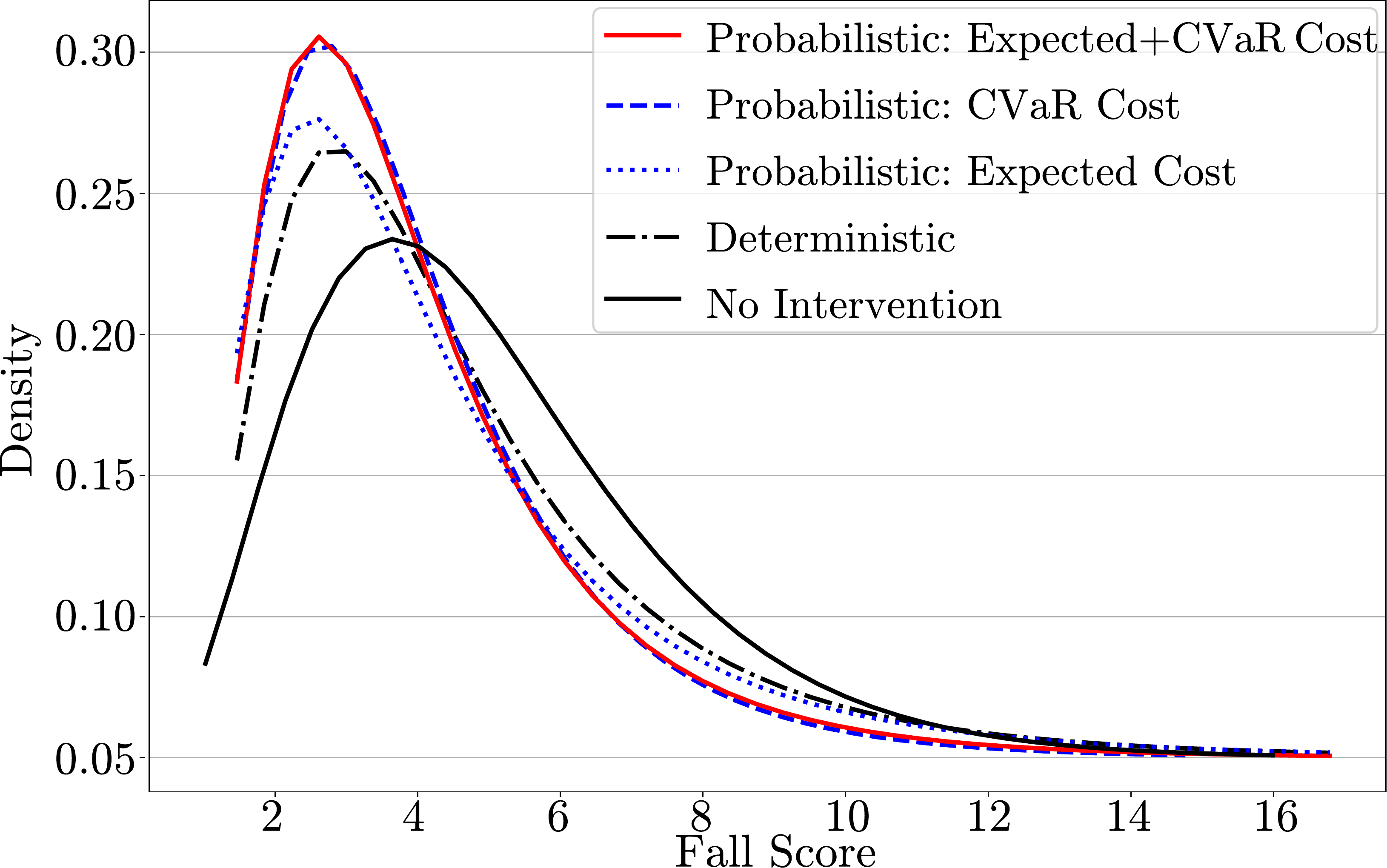}
\end{subfigure}
\caption{\small{Fall distribution from 20 scenarios in which the patient's initial pose is near the right side of the bed.}}
\label{fig:fall_distributions}
\vspace{-0.2in}
%\rule{3cm}{7cm}
\end{figure}

\begin{figure}[t!]
\centering
\begin{subfigure}
 \centering
 \includegraphics[width = 7cm]{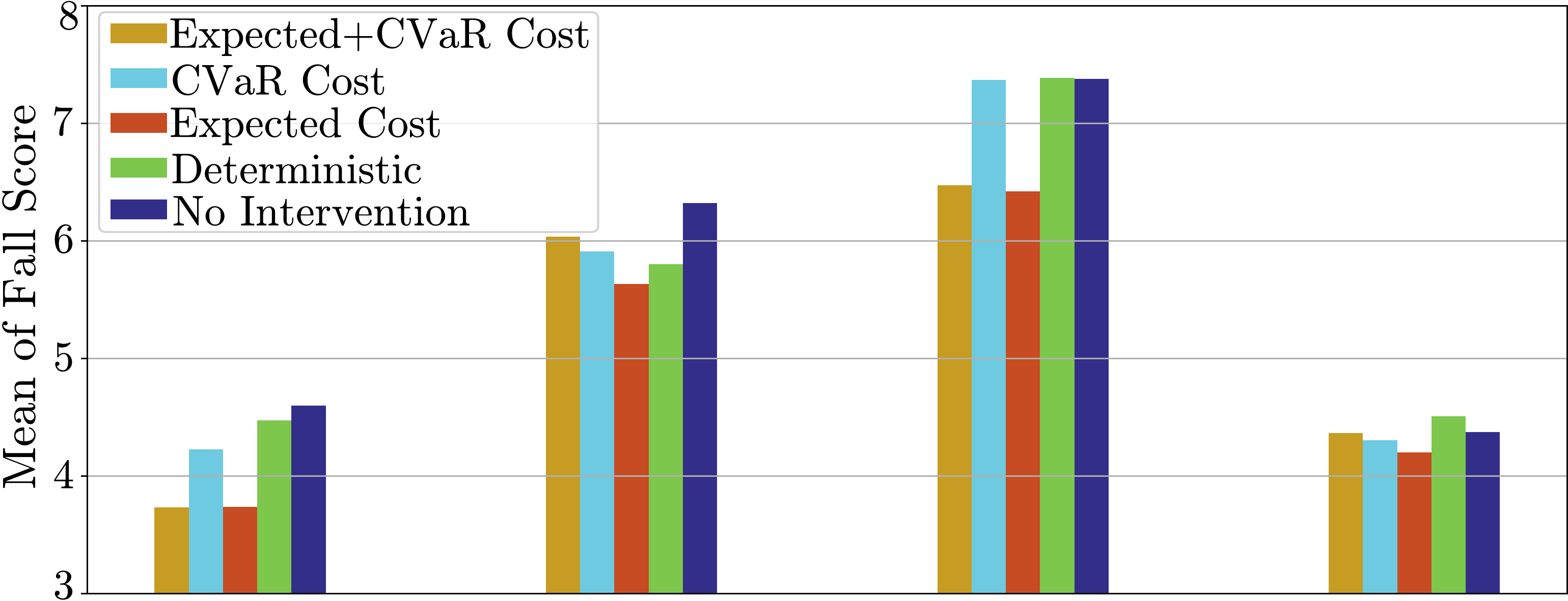}
\end{subfigure}
\begin{subfigure}
 \centering
\includegraphics[width = 7cm]{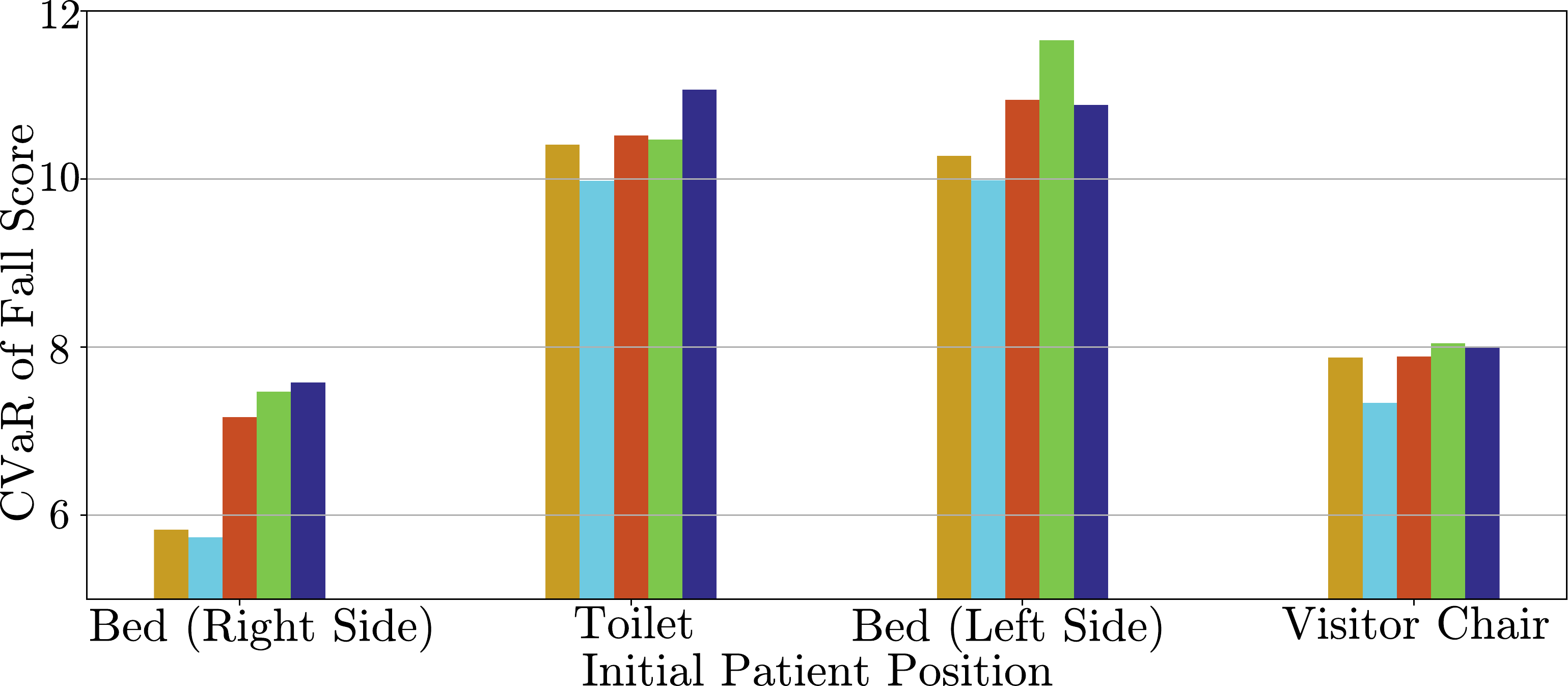} 
\end{subfigure}
\caption{\small{Mean and CVaR of fall scores distributions from 20 scenarios for four initial patient poses using different fall-prevention approaches. Overall, ``Expected+CVaR Cost" approach has a better performance considering these two metrics.  }}
\label{fig:fall_multi}
\vspace{-0.23in}
%\rule{3cm}{7cm}
\end{figure}

For each initial pose, we run 20 scenarios with random intentions~(sampled from a prior distribution over the intention set) and calculate the fall score along the actual patient path, before and after intervention. This results in a distribution of fall scores for a single initial pose which we show in Fig.\ \ref{fig:fall_distributions}. We provide the distributions for five different methods. We can see the effect of CVaR and expected cost functions on the mean and the area under the tail of the distribution. Using our probabilistic approach, we can reduce the number of rare events with high fall score. We also see that the ``CVaR" and ``expected + CVaR" cost functions perform best among all methods. Although the robot is successful in recognizing high fall score events even with low probability and takes action accordingly, the amount of change was not drastic due to the limited assistance of the walker. In our future studies, we would like to explore other assistance options.

Fig.~\ref{fig:fall_multi} provides the mean and CVaR of the fall score distributions from five methods for various initial poses. Here, we see that for some scenarios, different methods can result in similar performances. This is mainly in scenarios where the robot can provide the walker early in the patient's trajectory (such as scenarios where the patient starts from the visitor chair), resulting in the best performance possible for that scenario. However, there are some scenarios in which the robot must decide between multiple choices (such as the scenarios with the patient initial pose near the right side of the bed), creating the main challenge discussed in our problem. We see that, in those scenarios, our proposed cost function outperforms the other methods.

\begin{figure}[t!]
\centering
\includegraphics[width = 8.5cm]{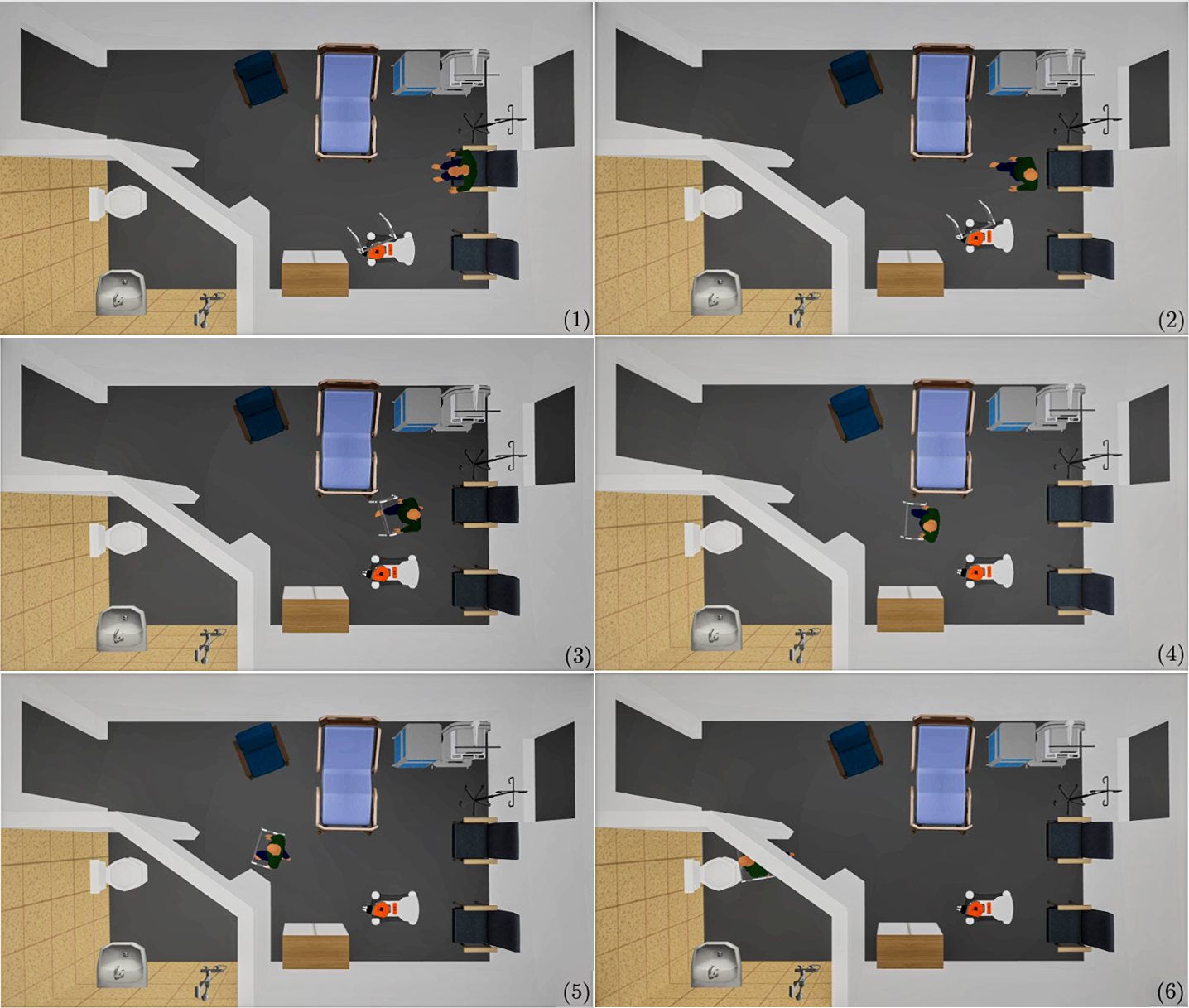}
\caption{\small{Snapshots of a risk-aware manipulation plan to deliver the walker to the patient, simulated in Gazebo.}}
\label{fig:frames}
\vspace{-0.23in}
%\rule{3cm}{7cm}
\end{figure}

Finally, in Fig.~\ref{fig:frames}, we present snapshots of a risk-aware plan to deliver a walker to the patient in the Gazebo simulation environment. At each time step, the robot predicts human motion and risk of fall based on the environmental factors and finds the best intervention point to minimize risk of fall.
%%%%%%%%%%%%%%%%%%%%%%%%%%%%%%%%%%%%%%%%%%%%%%%%%%%%%%5
\section{Conclusion}
\label{sec:conclusion}
We defined a new problem of patient fall prevention in hospitals through risk-aware planning for an assitive mobile robot. We formulated this as an optimization problem to reduce risk of fall by providing a mobility aid. 
Our approach combines learning-based human motion and intention prediction with model-based control. We proposed a new risk-sensitive cost function that considers the probability distribution of fall score over the entire predicted human motion distribution and avoids rare events of high fall score. We compared our proposed method with deterministic methods and showed that the combination of expected cost and CVaR risk metric works best in reducing the overall fall risk and avoiding high fall score events. Our proposed approach is limited to the problems where the human pose is known at each time step, and the room map~(including the positions of walls and objects) are known a priori. Moreover, this approach does not account for cases with other people moving in the environment. A local perception system for the robot could help with these issues.

There are several exciting options for the future work. First, we can extend our method to other types of assistive robots such as home robots. Using real data and cross-scene prediction methods, we can enable robots to assist in various settings. Another interesting addition to our work would be considering the human acceptance probability based on the object's configuration at the intervention point. A human acceptance model can simply be added into the cost function to increase the probability of the patient actually taking the walker when available. %Finally, it would be interesting to conduct human subject studies and investigate the long-term effects of an assistive robot equipped with our framework.

\section{Acknowledgments} 
We would like to thank Adam Conkey for providing the CEM solver used in this paper.
% \addtolength{\textheight}{-10cm}   % This command serves to balance the column lengths


\bibliographystyle{IEEEtran}
\bibliography{references}

\end{document}

\maketitle
\thispagestyle{empty}
\pagestyle{empty}

%%%%%%%%%%%%%%%%%%%%%%%%%%%%%%%%%%%%%%%%%%%%%%%%%%%%%%%%%%%%%%%%%%%%%%%%%%%%%%%%
\section{Human motion generation}
In this section, we provide more details on the optimization problem used for human motion generation. This provides a dataset for human motion and intention model learning. 

We assume that besides minimum path length and obstacle avoidance, patients try to avoid high fall risk situations too by moving closer to external support points. Assuming patients perform optimal motions with respect to path length and risk of fall, then we can formulate the trajectory prediction as the following optimization problem:
\begin{equation*}
\begin{aligned}
&\underset{\mathbf{\omega}}{\text{min}}
&&J = \sum_{t=1}^{h}(||\mathbf{\zeta}_t^P-\mathbf{\zeta}_g^P||_2^2 + \lambda d_t)&&\\
& \text{s.t.} & &f_{d} \leq 0&\\
&&& \zeta_t^P \not\in \mathcal{O}\quad &\forall t= 0,\dots,h\\
&&& d_t = \min(||\mathbf{\zeta}_t^P-\mathbf{\zeta}_i^{ESP}||_2^2) \quad &\forall t= 0,\dots,h\\
&&&&\forall i= 0,\dots,l
\end{aligned}
\end{equation*}

Where $\zeta_t^P$, $\zeta_i^{ESP}$ and $\zeta_g^P$ denote the patient's state at time $t$, position of available external support points and desired goal state, respectively. We solve the problem for $h$ time steps with $l$ external support points, and $m$ trajectory basis functions. $f_d$ encodes the dynamics constraints of the patient motion such as velocity and acceleration limits or the effect of obstacles on the patient's motion. We formulate obstacle avoidance in the second constraint. The last set of constraints is to find the closest external support point to each state of the patient along the horizon. This defines the distance to the closest supporting object which is used in the cost function to simulate the fragility of patients and the fact that they try to avoid falling by moving closer to supports along their trajectory.

\begin{figure}[t!]
\centering
\includegraphics[width = 7.7cm]{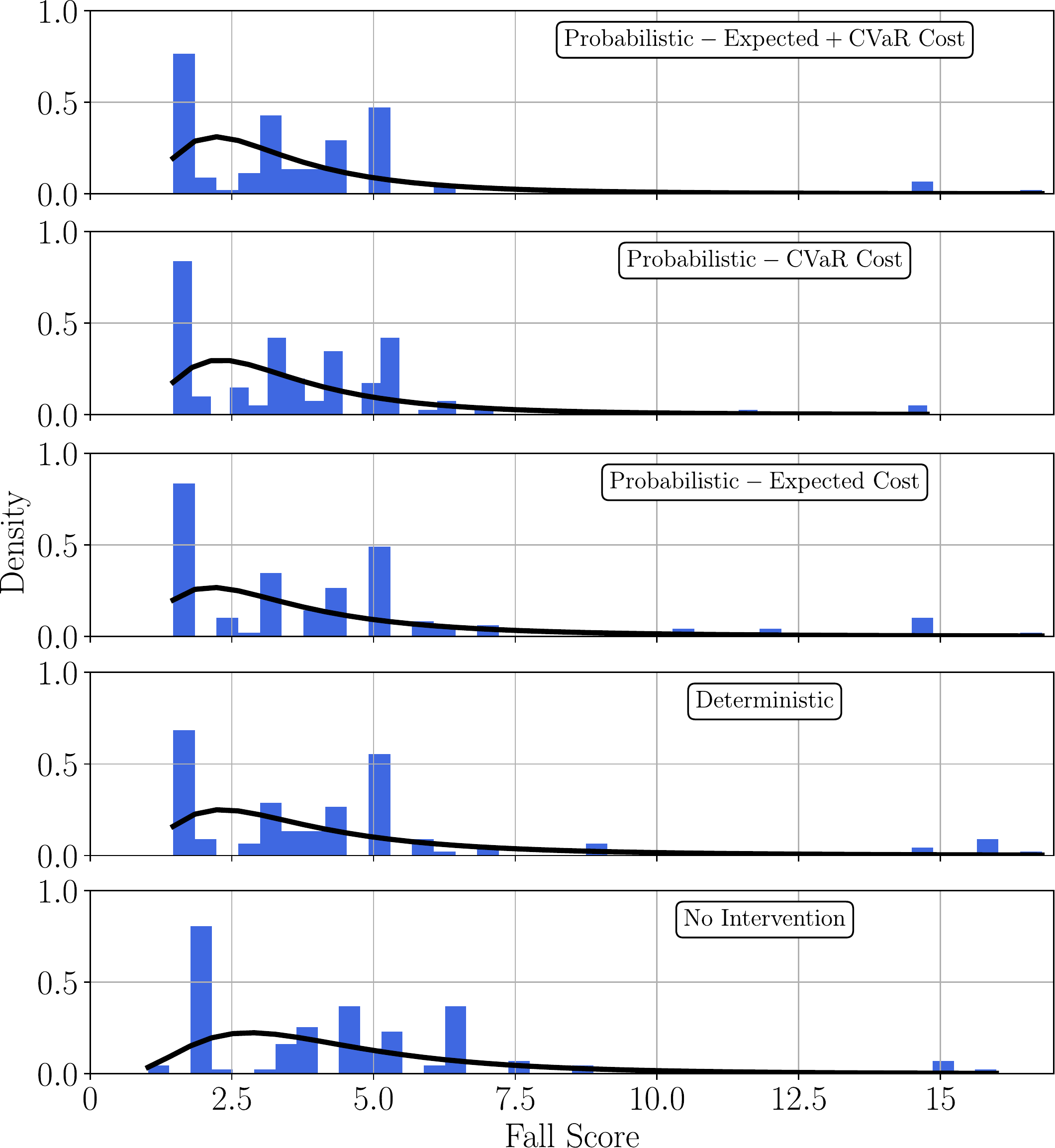}
\caption{\small{Fall distribution from 20 scenarios in which the patient's initial pose is near the right side of the bed.}}
\label{fig:fall_distribution_bed}
\vspace{-0.15in}
%\rule{3cm}{7cm}
\end{figure}

\begin{figure}[t!]
\centering
\includegraphics[width = 7.7cm]{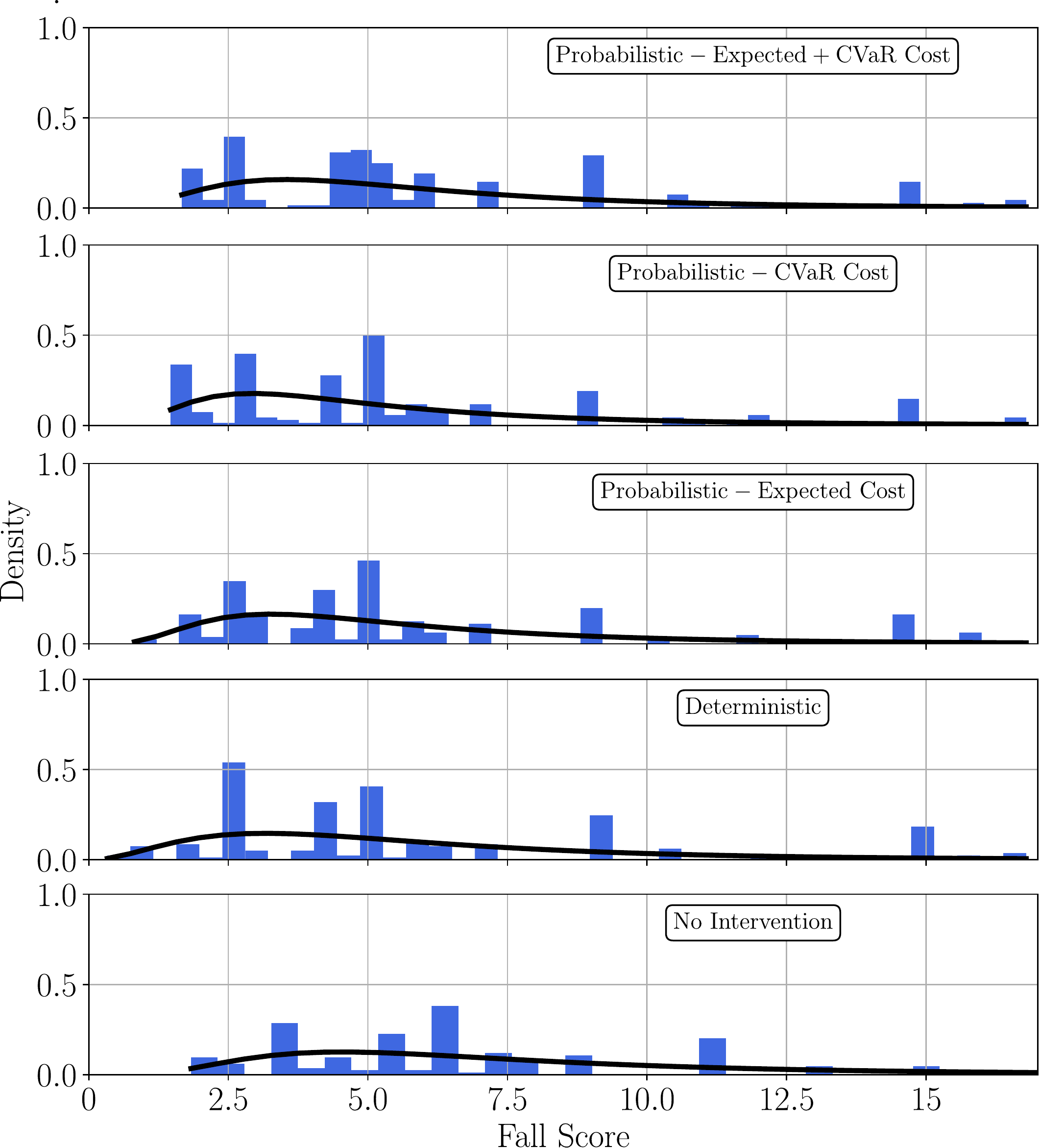}
\caption{\small{Fall distribution from 20 scenarios in which the patient's initial pose is near the toilet.}}
\label{fig:fall_distribution_door}
\vspace{-0.15in}
%\rule{3cm}{7cm}
\end{figure}

\begin{figure}[t!]
\centering
\includegraphics[width = 7.7cm]{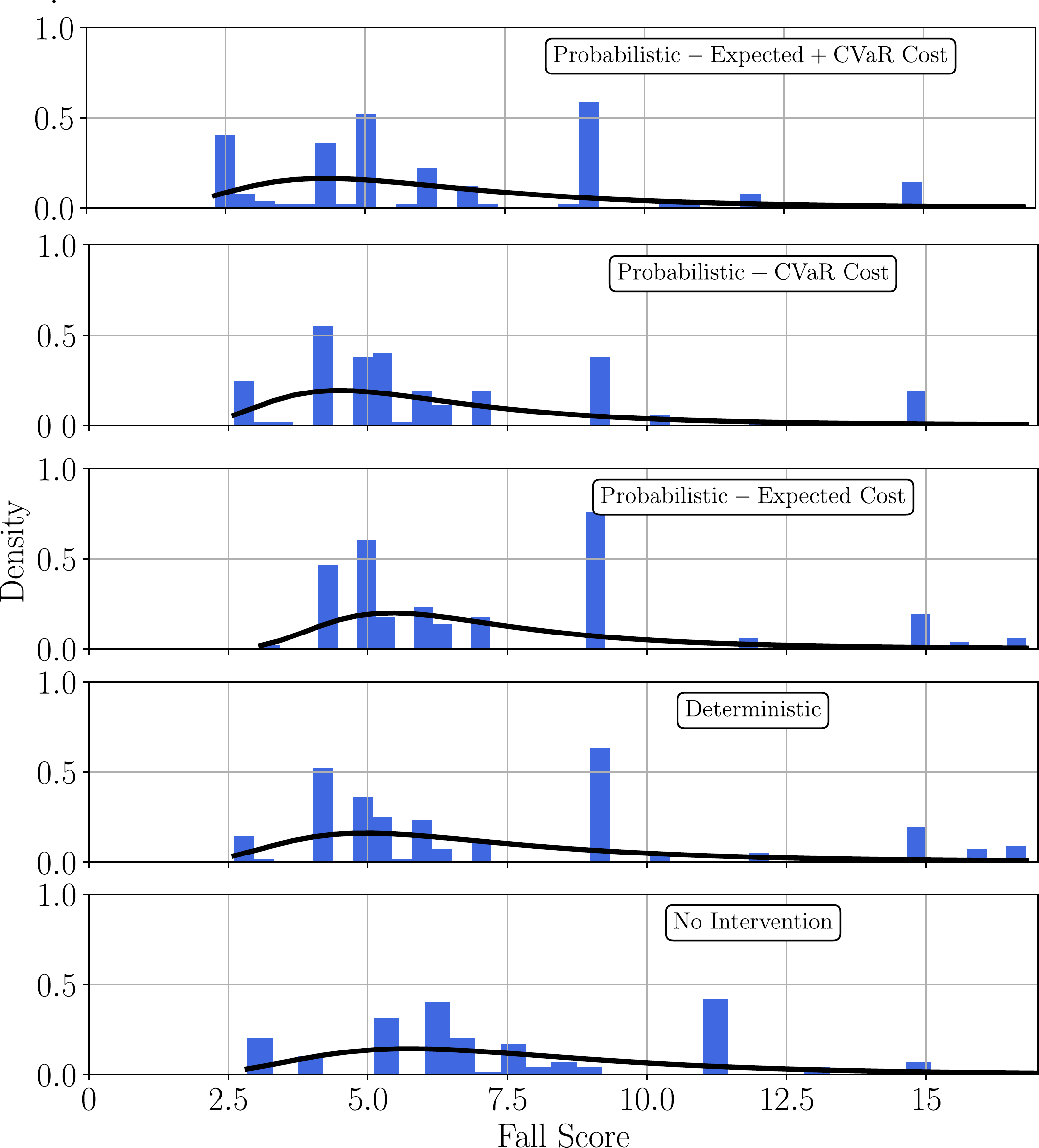}
\caption{\small{Fall distribution from 20 scenarios in which the patient's initial pose is near the left side of the bed.}}
\label{fig:fall_distribution_toilet}
\vspace{-0.15in}
%\rule{3cm}{7cm}
\end{figure}

\section{Results}

\begin{figure}[t!]
\centering
\includegraphics[width = 7.7cm]{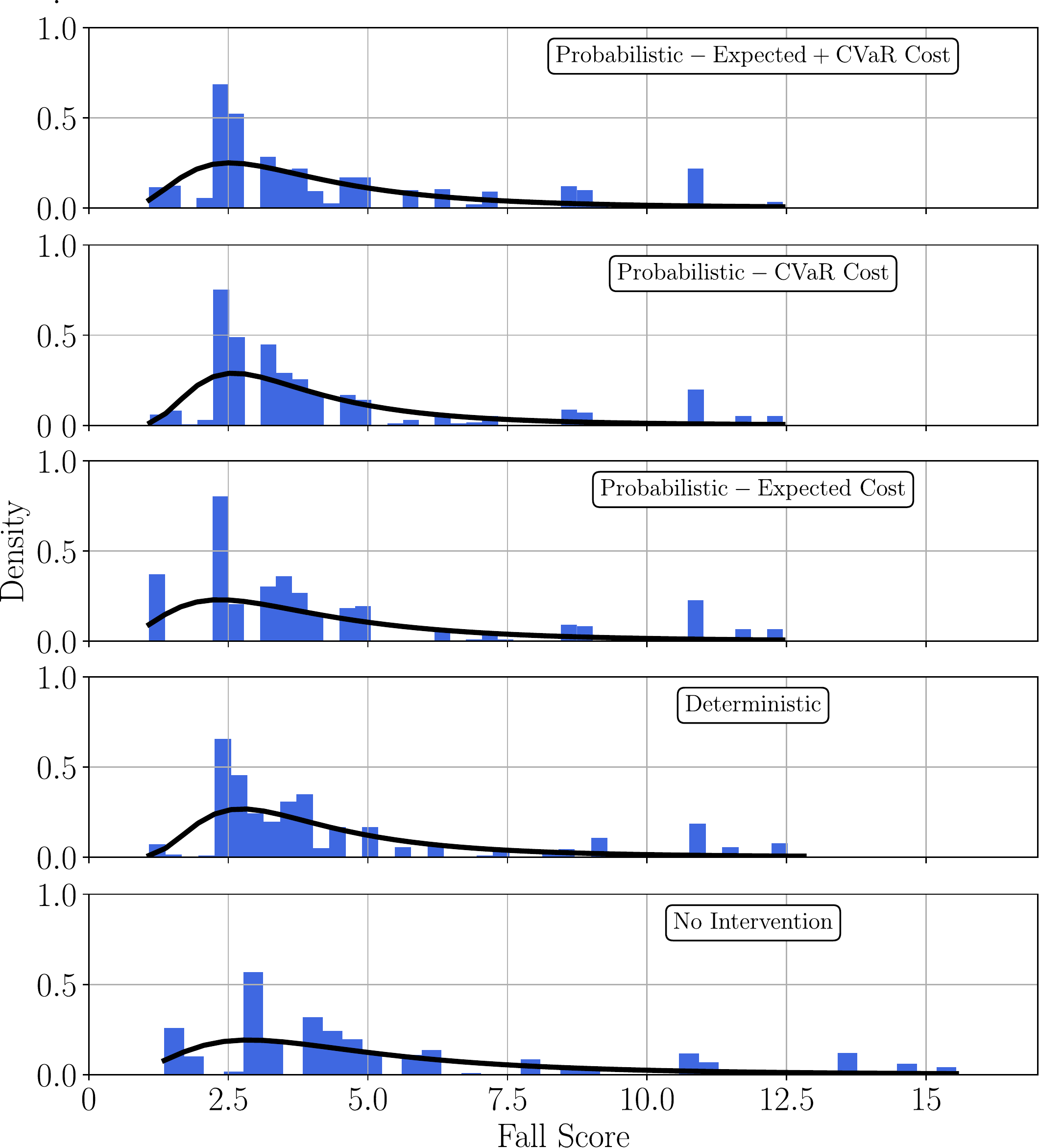}
\caption{\small{Fall distribution from 20 scenarios in which the patient's initial pose is near the visitor chair.}}
\label{fig:fall_distribution_chair}
\vspace{-0.15in}
%\rule{3cm}{7cm}
\end{figure}

Figs. \ref{fig:fall_distribution_bed}-\ref{fig:fall_distribution_chair} provide the fall score distributions from various experiments initiated from different locations within the hospital room. In each figure, we compare the results using different methods and cost functions. 

First, comparing the ``no-intervention" and ``deterministic" plots, we see the effect of utilizing a service robot to assist a hospital patient. Next, we can observe that using a probabilistic method has improved the overall performance of our planning considering a broad range of predictions. Finally, we can see that adding the CVaR risk metric has reduced the number of events with high fall score which is the main purpose of this research.

% \bibliographystyle{IEEEtran}

% \bibliography{references}